\documentclass[10pt, a4paper]{article}

\usepackage{lrec-coling2024} 

\usepackage{multirow}
\usepackage{paralist}
\usepackage{csquotes}
\usepackage{booktabs}

\title{On Zero-Shot Counterspeech Generation by LLMs}

\name{Punyajoy Saha\textsuperscript{\rm 1}, Aalok Agrawal\textsuperscript{\rm 1}, Abhik Jana\textsuperscript{\rm 2}, Chris Biemann\textsuperscript{\rm 3} , Animesh Mukherjee\textsuperscript{\rm 1}} 

\address{
\textsuperscript{\rm 1} Indian Institute of Technology, Kharagpur, \textsuperscript{\rm 2} Indian Institute of Technology, Bhubaneswar \\
\textsuperscript{\rm 3} Universität Hamburg, Germany \\
         punyajoys@iitkgp.ac.in,
         aalokagrawal@iitkgp.ac.in,
         abhikjana@iitbbs.ac.in,\\
        biemann@informatik.uni-hamburg.de,
    animeshm@cse.iitkgp.ac.in\\}

\abstract{
With the emergence of numerous Large Language Models (LLM), the usage of such models in various Natural Language Processing (NLP) applications is increasing extensively. Counterspeech generation is one such key task where efforts are made to develop generative models by fine-tuning LLMs with hatespeech - counterspeech pairs, but none of these attempts explores the intrinsic properties of large language models in zero-shot settings. 
In this work, we present a comprehensive analysis of the performances of four LLMs namely GPT-2, DialoGPT, ChatGPT and FlanT5 in zero-shot settings for counterspeech generation, which is the first of its kind. For GPT-2 and DialoGPT, we further investigate the deviation in performance with respect to the sizes (small, medium, large) of the models. On the other hand, we propose three different prompting strategies for generating different types of counterspeech and analyse the impact of such strategies on the performance of the models. Our analysis shows that there is an improvement in generation quality for two datasets (17\%), however the toxicity increase (25\%) with increase in model size. Considering type of model, GPT-2 and FlanT5 models are significantly better in terms of counterspeech quality but also have high toxicity as compared to DialoGPT.  ChatGPT are much better at generating counter speech than other models across all metrics. In terms of prompting, we find that our proposed strategies help in improving counter speech generation across all the models. \\ \newline  }

\begin{document}

\maketitleabstract

\section{Introduction}
Large Language Models (LLMs) like GPT-3, BARD, LLaMA are being used to produce state-of-the-art performances for numerous NLP tasks, e.g., summarisation, machine translation, text classification, etc. Despite the promising capabilities of LLMs, researchers point out their limitations in certain genres of NLP tasks like question answering~\cite{zheng2023does}. Digging deep into the LLMs' behaviour for a spectrum of NLP tasks provide insights about their intrinsic properties and usabilities for such tasks. Therefore, in this paper, we investigate the applicability and limitations of LLMs for one of the key NLP task of \textit{counterspeech} generation.

The rise of social media and online platforms has provided individuals with unprecedented opportunities to express themselves and engage in discussions on a global scale. Often these expressions become toxic due to bad actors spreading hate speech\footnote{\url{https://help.twitter.com/en/rules-and-policies/hateful-conduct-policy}}.
To curtail hate speech proliferation, the moderation community has come up with the strategy 
of producing extensive \textit{counterspeech} which is a direct response countering the hate/abusive speech. An example is shown below.

\begin{displayquote}
{\footnotesize{
\noindent\textbf{Hate speech}: \textit{Jews cannot be patriots, since their allegiance will always be to the state of Israel.}

\noindent\textbf{Counterspeech}: \textit{You can have parents and grandparents born elsewhere and still be a patriot for the country you were born in.}}} \\
\end{displayquote}
Recently, the NLP community has started exploring the usefulness of LLMs for the task of vanilla counterspeech generation as well as categorical counterspeech generation. While we find several works in the line of finetuning LLMs with hate speech - counterspeech pairs~\cite{zhu-bhat-2021-generate}, adding additional context to the LLMs~\cite{li-etal-2022-knowledge}, none of them study the intrinsic properties of these models or explore prompting in a zero-shot setting to generate a specific type of counterspeech.

\noindent In this paper, our contributions can be summarized as follows.
\begin{compactitem}
    \item We investigate the applicability of four LLMs (GPT-2, DialoGPT, FlanT5 and ChatGPT) for zero-shot counterspeech generation, which is the first ever attempt of this kind. We evaluate these models over four different counterspeech generation dataset -- CONAN~\cite{chung2019conan}, CONAN-MT~\cite{fanton2021human}, Reddit and Gab~\cite{qian2019benchmark}. We compare and analyse these LLMs' performances to come up with insightful observations which could be useful for the research community. We further dig into variations of a particular model in terms of size and analyse if that has any effect in counterspeech generation.
    \item We propose three prompting strategies, namely \textit{manual}, \textit{frequency based}, and \textit{cluster centered}. We analyze the effect of these strategies on categorical counterspeech generation.
\end{compactitem}

\noindent From our detailed analysis, we make the following key observations.

\noindent\textbf{Overall performance}: ChatGPT outperforms all other models (DialogGPT, FlanT5, GPT-2 along with their variants) in terms of generational metrics - gleu (12\%), meteor (32\%) and bleurt (42.25\%). On the other hand, counterspeech quality and argument quality improves by 120\% and 35\% respectively. One concerning observation is that the readability of the chatGPT generated posts reduces by 35\%. \\
\noindent\textbf{Effect of model size}: With the increase of the size of the models (both for DialogGPT and GPT-2), there is an increase in \textit{toxicity} in responses by 44\%, 25\%, and 30\% for CONAN-MT, Reddit, and Gab respectively. \\

\noindent\textbf{Effect of prompt type}: Manual prompts perform better across \textit{denouncing}, \textit{facts} and \textit{humour} type counterspeech. Cluster-centered prompts are better for \textit{affiliation} type counterspeech for GPT2 and DialogGPT, while for the same type, frequency based prompts are better for FlanT5 and ChatGPT.
    
We make our code and resource 
used for this research public~\footnote{\url{https://github.com/aalokagrawal/Zeroshot_Counterspeech}} for reproducibility purposes.

\section{Related work}

\noindent\textbf{Large language models}: A language model estimates the probability distribution over a text. Recent advancement in the scaling of such models from a few million parameters~\cite{meritypointer} to hundred million parameters~\cite{brown2020language} and larger dataset~\cite{gao2020pile} have made them \textit{large} language models (LLMs) performing better at a lot of downstream tasks. At this scale, the model can easily learn the downstream tasks in the few-shot as well as the zero-shot setting~\cite{radford2019language}. In the zero-shot setting, we only provide template (prompts) to the models where the prompts are selected using prompt engineering.\\
\noindent\textbf{Prompt engineering}: Prompt based learning methods learn LMs that compute the probability $P(\textbf{x};\theta)$ of the prompt text \textbf{x} itself and use this probability to predict the next sequence \textbf{y}~\cite{liu2023pre}. So, designing the template for \textbf{x} is an important step when these to compute the LM probabilities. Methods for text generation tasks in zero-shot setting involve manually adding prompts to perform summarization and machine translation tasks~\cite{radford2019language}. Recent methods have involved finding trigger tokens which can be used to create the prompt and do the task in zero-shot settings~\cite{shin-etal-2020-autoprompt}.
\noindent\textbf{Counterspeech generation}: An effective strategy to mitigate hate speech is counterspeech as it does not violate the freedom of expression~\cite{benesch2016considerations}. While the idea of countering some hateful messages is not new, the research community has recently started taking a huge interest in understanding counterspeech practices and their effectiveness while mitigating hate speech~\cite{mathew2019thou}. Recently, \citeauthor{tekirouglu2020generating} proposed novel techniques to generate counterspeech using a GPT-2 model with post-facto editing by experts or annotator groups. One of the recent generation methods uses a three-stage pipeline -- \textit{Generate, Prune} and \textit{Select} (GPS) to generate diverse and relevant counterspeech output~\cite{zhu2021generate}. Another work focuses on adding knowledge grounding to the generation pipeline~\cite{chung-etal-2021-towards}. A recent work further analyses several such language models and decoding strategies after finetuning them~\cite{tekirouglu2022using}.  
Past research has highlighted the usage of different types of counterspeech~\cite{benesch2014countering}. Two of the past works further built dataset and models for type specific counterspeech~\cite{mathew2019thou,chung2021multilingual}. Only one work focuses on generation of type specific counterspeech using generative discriminator (GEDI) models~\cite{ijcai2022p716}.

In all the past studies, they use a hate speech-counterspeech dataset to finetune the models first and then show their evaluation. On the other hand, our objective was to understand what these models know intrinsically. Hence, we study these LLMs in zero-shot setting to understand their intrinsic capability and compare them in terms of size and types. Second, we propose several prompting strategies which can generate type specific counterspeech as well in the zero-shot setting and further compare the different LLMs. 

\section{Datasets}

\subsection{Hatespeech-Counterspeech datasets}
In order to evaluate our approach, we use four public datasets which contain hate speech and its corresponding counterspeech. The details of these datasets are noted in Table \ref{tab:cs_dataset}. Reddit and Gab datasets contain $5,257$ and $14,614$ hate speech instances, respectively~\cite{qian2019benchmark}. We use the English part of the CONAN dataset~\cite{chung2019conan} which contains $408$ hate speech instances. The multitarget CONAN dataset~\cite{fanton2021human} contains $3,718$ hate speech instances. The counterspeech in Gab and Reddit datasets was written by AMT workers, whereas for CONAN, the counterspeech was written by expert NGO operators. For the CONAN-MT dataset, the pairs were generated by a generation model and later reviewed by experts.

We further made hate speech and counterspeech pairs from these datasets, such that each hate speech was associated with one counterspeech. Finally, we ended up with 3,864, 14,223, 41,580, and 5003 datapoints for CONAN, Reddit, Gab, and CONAN-MT, respectively.irst, we need to divide the dataset into train and test splits. For the smaller datasets - CONAN and CONAN-MT, we find all the unique hatespeech and randomly split into  train and test set with 80\% for training and 20\% for test set. For larger datasets - Gab and Reddit, we randomly take 500 hatespeech samples for the set of hatespeech samples. We make sure the hate speech instances in the train/validation set are not repeated in the test set to evaluate in the wild setting. For all our experiments, we use the test part of these datasets.

\begin{table}[!htpb]
\centering
\scriptsize
\begin{tabular}{l|l|l|l|l|l}\toprule
\textbf{Dataset} & \textbf{Source-H} & \textbf{Source-C} & \textbf{Hate ins} & \textbf{\# pairs} & \textbf{Avg. len} \\ \midrule
CONAN-MT    & synthetic     & expert     &   3,718     &    5,003 & 25.15    \\ 
CONAN       & synthetic     & expert     &   408     &   3,864   & 24.07   \\ 
Reddit      & reddit        & crowd      &   5,257    &   14,223   & 15.55  \\ 
Gab         & gab           & crowd      &   14,614    &   41,580   & 16.05  \\ \bottomrule
\end{tabular}
\caption{\footnotesize This table presents the source of hate speech (Source-H), source of counterspeech (Source-C), hate speech instances (Hate ins), the total pairs (\# pairs) and average length of the counterspeech (Avg. len) for each of the CONAN, CONAN-MT, Reddit and Gab dataset.}
\label{tab:cs_dataset}
\end{table}

\subsection{Addtional datasets}
\label{sec:additional}
In this section, we describe the additional datasets required for the evaluation of the generated responses. 

\noindent\textbf{Counterspeech dataset}: For measuring the quality of counterspeech, we use two datasets from \citet{mathew2018analyzing} and \citet{chung2021multilingual}. Both these datasets have type information, while \citet{mathew2019thou} in addition have non-counterspeech comments curated from YouTube. We use two variations of the counterspeech dataset. 

The first variant compiles the counterspeech posts themselves. This is primarily used for \textbf{classification} of a text into a counterspeech or not. In order to align our settings with the recommendations given by ~\citet{chung2021towards}, we place the hostile category counterspeech in the non-counterspeech part in this variant~\cite{mathew2019thou}. This way, we had 4,175 counterspeech comments and 9,765 non-counterspeech comments. We divide the dataset into train:validation:test in the ratio of 8:1:1 using stratified sampling. 

The second variant compiles the counterspeech types. This is primarily used for \textbf{classification} of a counterspeech into one of its types and \textbf{prompting}. We take the counterspeech posts from \citet{mathew2019thou} and merge them with the counterspeech from ~\citet{chung2021multilingual} along with their types. One thing to note is that we don't utilise the posts from \citet{mathew2019thou} which contain more than one strategies of counterspeech. Next, we study the definition of different types of counterspeech and select six types of counterspeech which appeared distinctive to all the authors unanimously. The statistics of the dataset finally extracted is noted in Table~\ref{tab:cs_types}. 

We divide the dataset, with 30\% of it for constructing \textbf{prompts} and the rest for classification using stratified sampling. The classification dataset was further divided into train:validation:test in the ratio of 8:1:1 using stratified sampling. The prompt dataset is used to create the type prompts. 

\begin{table}[!htpb]
\centering
\scriptsize
\begin{tabular}{|l|l|l|l|l|l|}
\toprule
\textbf{Type} &\textbf{Classification} & \textbf{Prompting} & \textbf{F1} & \textbf{clusters} \\ \midrule
hypocrisy    & 579  & 248 & 0.59  & 15\\
denouncing   & 738  & 316 & 0.85  & 16 \\ 
humor        & 607  & 260 & 0.76  & 16\\ 
facts        & 1094 & 469 & 0.84  & 18 \\ 
affiliation  & 163  & 70  & 0.84  & 16 \\
question     & 227  & 97  & 0.97  & 18\\ \midrule 
average/total& 3408 & 1460 & 0.80 & 16.5\\ \bottomrule
\end{tabular}
\caption{\footnotesize  This table represents the type specific information for each of the type of counterspeech that we considered for our task. The columns \textit{Prompting} and \textit{Classification} represents the amount of data points used for finding prompt strategies and building classification model. \textit{F1 score} (F1) column shows the performance of the type classifier. \textit{Clusters} column represents the number of clusters found per type using the cluster-centered prompting strategy.}
\label{tab:cs_types}
\end{table}

\noindent\textbf{Counterargument dataset}: For evaluating the counterargument quality we select a popular argument dataset~\cite{stab-etal-2018-cross} which has 6,317 against and 4,822 for arguments categorized into six topics. For each topic, we assume all possible pairs of arguments. From this set, we sample (without replacement) 10,000 pairs which have the same stance and 10,000 pairs which have the opposite stance. This way, we have a dataset of 60,000 argument pairs. We divide the dataset into train:validation:test in the ratio of 8:1:1 using stratified splitting.

\section{Methodology}

\subsection{Models}

We use three different model variants for understanding the zero-shot capability in counterspeech generation.

\noindent\textbf{GPT-2}~\cite{radford2019language} is trained on a large dataset called WebText. The dataset contains the textual content found in 45 million links shared by users on Reddit~\cite{trinh2018simple}. Note that, WebText is not directly sourced from Reddit itself, but rather consists of data derived from outbound links posted on Reddit. This language model is trained with the objective of predicting the next word, given all the previous words within some text. The model aims at maximising $p(x) = \prod^n_{i=1} p(x_n|x,...,x_{n-1})$, for our experiments we use all three different versions of GPT-2 – 117M (\textbf{s}mall), 345M (\textbf{m}edium), and 762M (\textbf{l}arge) parameters from this link\footnote{https://huggingface.co/docs/transformers/model\_doc/gpt2}.

\noindent\textbf{DialoGPT}~\cite{zhang2020dialogpt} is trained on a large corpus consisting of English Reddit dialogues. The corpus consists of 147 million instances of dialogues, collected over a period of 12 years. Unlike GPT-2, this model generates better dialogue-like responses to any given prompt. In this model, along with ground truth response $T=x_1,...,x_n$, we also have a dialogue utterance history $S$. The model aims at maximising $p(T|S) = p(x_1|S)\prod^n_{i=2} p(x_i|S,...,x_{i-1})$. For our experiment, we use all three different versions of DialoGPT - 117M (\textbf{s}mall), 345M (\textbf{m}edium), and 762M (\textbf{l}arge) parameters from this link\footnote{https://github.com/microsoft/DialoGPT}.

\noindent\textbf{FlanT5}~\cite{chung2022scaling} is a T5~\cite{JMLR:v21:20-074} model that has been finetuned on a multi-task mixture of supervised tasks and for which each task is converted into a text-to-text format. They use instruction finetuning~\cite{ouyang2022training} procedure on each of these tasks, as well as chain-of-thought (CoT) prompting~\cite{wei2022chain}. Overall, the authors show that using such a framework improves the results across various benchmarks over the T5 versions.

\noindent\textbf{ChatGPT}~\cite{Introduc70:online} is trained with a GPT 3.5 model using Reinforcement Learning from Human Feedback (RLHF), using a similar method as InstructGPT~\cite{ouyang2022training} but with slight differences in the data collection setup. ChatGPT performs exceptionally well in question-answering scenarios. Beyond its capability of being a conversational tool, many attempts have been made to evaluate the quality of ChatGPT-generated texts in various domains~\cite{yang2023harnessing}.

\subsection{Prompting strategies}
\label{sec:prompt}

Type specific generation is another challenge in counterspeech generation~\cite{mathew2019thou}. In this regard, the first few words of the counterspeech can be essential for the type we want to generate. We propose three different prompting strategies to generate the first few words (prompts). These prompts help us in controlling the type of the counterspeech generated by the above LLMs.

\noindent\textbf{Manual prompting}: In this strategy, two authors experienced in hate speech detection research read through the prompts dataset and created 2-3 possible beginnings (prompts) for each type of counterspeech. These prompts guide the model to generate the appropriate type of counterspeech. For manual prompts, we do not set a hard limit on the number of words but asked the authors to make them small.

\noindent\textbf{Frequency based prompting}: In this strategy, we collect the beginning four words as a sub-string from each counterspeech of each type and cluster using exact matching of the sub-strings. These sub-strings represent the prompts for a particular type. We take the top five prompts based on their frequency for each type. 

\noindent\textbf{Cluster centered prompting}: In this strategy, we first pass all the counterspeech from the prompting dataset through sentence embedding model \texttt{all-mpnet-base-v2}\footnote{https://huggingface.co/sentence-transformers/all-mpnet-base-v2}.  For each type of counterspeech we then cluster the embeddings using $K$-means clustering. We decide the number of cluster for each type of counterspeech using the elbow method~\cite{8549751}. We note the number of clusters for each type in \ref{tab:cs_types}. The average clusters are $\sim 16$ per type. Out of these clusters we select the top 10 clusters using their cluster size. Then we select top three sentences per cluster which lie closest to the cluster center. These sentences act as a representative of that cluster. The final prompts per cluster are comprising the beginning 4 words as a sub-string from each of these three sentences. This way we collect 30 prompts per type in total.

We note one instance of prompts for each strategy per type in the Table \ref{tab:prompt_types}. 

\begin{table}[!htpb]
\scriptsize
\centering
\begin{tabular}{l|l|l|l}
\toprule
\textbf{Type} & \textbf{M} &\textbf{CC} & \textbf{FB}\\ \midrule
F     & This is a fact & the myth that muslims & The vast majority of\\
Hy & In contradiction & i am wondering have & If you are really \\
Hu    & This is funny  & i bet she got & Must be hard for \\
A & I also belong  &  i am jewish and & I am a christian\\
Q & Are you aware of & how do you know & Why do you think\\
D & Please do not say & why is the hate & Why is this a\\\bottomrule
\end{tabular}
\caption{\footnotesize  One instance of prompt using each prompt strategy per type. Each column represents the different prompt strategies and each row represents a particular type of counterspeech. M: Manual, CC: Cluster centered, FB: Frequency based. F: Facts, Hy: Hypocrisy, Hu: Humour, Af: Affiliation, Q: Questions, D: Denouncing.}
\label{tab:prompt_types}
\end{table}

\subsection{Experimental setup}

Our experimental setup comprises generation of a post when we pass the hate speech as an input to the LLMs. In case of type prompts, we further add the \texttt{type-prompt} at the end of the hate speech as input. For ChatGPT API, we use a similar setting where we just pass the hate speech without any prompts to the ChatGPT API; for the type prompts we add ``start the counterspeech with following \texttt{type-prompt}". We extract the \texttt{type-prompt} by randomly selecting a prompt from the given list of prompts for a particular type.

Sequences are generated up to a minimum length of $40$ and a maximum of $60$ tokens. We set the generation length a bit higher as compared to average length of the counterspeech (as noted in Table \ref{tab:cs_dataset}) in order to allow the model more freedom to generate. In addition, for DialoGPT, FlanT5 and GPT-2 models, top $k$ sampling and top $p$ sampling (aka nucleus sampling) are used while generating from our trained models. At each generation step, all the generated tokens are ranked according to their probabilities, and the top 100 most probable tokens are selected for broad distributions. In the case of narrow distributions, all the tokens are included until their CDF is 0.92 following the recommendations by~\citet{holtzman2019curious}. The temperature is 1.2, and the repetition penalty is 3.5. For ChatGPT API, we add a system message ``you are a helpful assistant that generates counterspeech''. We also keep top $p$ sampling, repetition penalty and the temperature same.

\begin{table*}
\scriptsize
\centering
\begin{tabular}{l|c|c|c|c|c|c|c|c|c|c}
\toprule
\textbf{model} & \textbf{gleu} & \textbf{met} & \textbf{div} & \textbf{nov} & \textbf{blrt} & \textbf{cs} & \textbf{c\_arg} & \textbf{arg} & \textbf{tox ($\downarrow$)} & \textbf{fre} \\\midrule
\multicolumn{11}{c}{\textbf{CONAN\_MT}} \\\midrule
DGPT-(s) & 0.07 & 0.08 & \textbf{0.84} & 0.84 & -1.13 & 0.15 & 0.54 & 0.26 & 0.24 & 65.21 \\
DGPT-(m) & 0.07 & 0.08 & 0.84 & 0.84 & -1.16 & 0.16 & 0.54 & 0.22 & 0.19 & \textbf{72.46} \\
DGPT-(l) & 0.07 & 0.09 & 0.83 & 0.83 & -1.14 & 0.15 & 0.59 & 0.23 & 0.28 & 66.07 \\
GPT-2     & 0.06 & 0.11 & 0.82 & 0.84 & -1.02 & 0.56 & 0.50 & 0.38 & \textbf{0.13} & 43.24 \\
GPT-2-(m) & 0.06 & 0.10 & 0.83 & \textbf{0.85} & -1.05 & 0.51 & 0.49 & 0.36 & 0.18 & 44.25 \\
GPT-2-(l) & 0.06 & 0.11 & 0.83 & 0.84 & -1.04 & 0.48 & 0.47 & 0.36 & 0.19 & 43.27 \\
flan-T5-(s) & 0.08 & 0.13 & 0.81 & 0.82 & -0.94 & 0.40 & 0.56 & 0.48 & 0.18 & 61.21\\
flan-T5-(b) & 0.08 & 0.13 & 0.81 & 0.81 & -0.90 & 0.43 & 0.49 & 0.47 & 0.21 & 60.65\\
flan-T5-(l) & 0.08 & 0.12 & 0.82 & 0.82 & -0.96 & 0.41 & 0.46 & 0.43 & 0.18 & 58.08\\
ChatGPT  & \textbf{0.09} & \textbf{0.17} & 0.66 & 0.80 & \textbf{-0.53} & \textbf{0.95} & \textbf{0.64} & \textbf{0.51} & 0.15 & 29.89 \\\midrule
\multicolumn{11}{c}{\textbf{CONAN}} \\\midrule
DGPT-(s) & 0.09 & 0.11 & \textbf{0.88} & \textbf{0.87} & -1.21 & 0.15 & 0.58 & 0.20 & 0.31 & 60.67 \\
DGPT-(m) & 0.09 & 0.11 & \textbf{0.88} & \textbf{0.87} & -1.23 & 0.09 & 0.54 & 0.15 & 0.24 & \textbf{70.95} \\
DGPT-(l) & 0.09 & 0.12 & 0.86 & 0.86 & -1.20 & 0.08 & 0.65 & 0.21 & 0.37 & 63.57 \\
GPT-2     & 0.08 & 0.15 & 0.85 & 0.86 & -1.06 & 0.48 & 0.55 & 0.37 & \textbf{0.20} & 41.09 \\
GPT-2-(m) & 0.08 & 0.15 & 0.85 & 0.86 & -1.06 & 0.34 & 0.54 & 0.38 & 0.23 & 44.65 \\
GPT-2-(l) & 0.08 & 0.15 & 0.85 & 0.86 & -1.08 & 0.43 & 0.51 & 0.35 & 0.21 & 43.73 \\
Flan-T5-(s) & 0.10 & 0.17 & 0.84 & 0.84 & -1.86 & 0.33 & 0.56 & 0.40 & 0.26 & 61.59\\
Flan-T5-(b) & 0.10 & 0.17 & 0.84 & 0.84 & -1.84 & 0.33 & 0.52 & 0.44 & 0.25 & 53.48\\
Flan-T5-(l) & 0.10 & 0.17 & 0.84 & 0.84 & -0.98 & 0.31 & 0.58 & 0.42 & 0.22 & 55.32\\
ChatGPT  & \textbf{0.12} & \textbf{0.23} & 0.69 & 0.81 & \textbf{-0.63} & \textbf{0.89} & \textbf{0.64} & \textbf{0.44} & 0.23 & 32.05 \\\hline
\multicolumn{11}{c}{\textbf{Gab}} \\\midrule
DGPT-(s) & 0.05 & 0.07 & \textbf{0.87} & \textbf{0.86} & -1.26 & 0.06 & 0.53 & 0.06 & \textbf{0.09} & 58.18 \\
DGPT-(m) & 0.05 & 0.07 & 0.86 & \textbf{0.86} & -1.26 & 0.08 & 0.55 & 0.06 & \textbf{0.09} & 57.00 \\
DGPT-(l) & 0.05 & 0.09 & 0.85 & 0.84 & -1.28 & 0.07 & \textbf{0.56} & 0.06 & \textbf{0.09} & 59.25 \\
GPT-2     & 0.05 & 0.12 & 0.83 & 0.85 & -1.37 & 0.31 & 0.53 & 0.19 & 0.15 & 58.16 \\
GPT-2-(m) & 0.05 & 0.12 & 0.84 & 0.85 & -1.37 & 0.28 & 0.54 & 0.19 & 0.19 & 58.47 \\
GPT-2-(l) & 0.05 & 0.12 & 0.83 & 0.85 & -1.36 & 0.28 & 0.53 & 0.19 & 0.19 & 55.44 \\
FlanT5-(s) & 0.06 & 0.11 & 0.84 & 0.84 & -1.37 & 0.24 & \textbf{0.56} & 0.22 & 0.16 & 67.10\\
FlanT5-(b) & 0.06 & 0.11 & 0.84 & 0.83 & -1.35 & 0.23 & 0.50 & 0.21 & 0.20 & \textbf{68.33}\\
FlanT5-(l) & 0.06 & 0.11 & 0.84 & 0.83 & -1.34 & 0.26 & 0.52 & 0.19 & 0.16 & 63.79\\
ChatGPT  & \textbf{0.08} & \textbf{0.17} & 0.64 & 0.80 & \textbf{-0.71} & \textbf{0.90} & 0.46 & \textbf{0.26} & 0.12 & 29.77 \\\midrule
\multicolumn{11}{c}{\textbf{Reddit}} \\\midrule
DGPT-(s) & 0.05 & 0.06 & \textbf{0.87} & \textbf{0.88} & -1.22 & 0.07 & 0.59 & 0.07 & \textbf{0.07} & 30.52 \\
DGPT-(m) & 0.05 & 0.07 & \textbf{0.87} & 0.87 & -1.21 & 0.08 & 0.55 & 0.06 & \textbf{0.07} & 58.24 \\
DGPT-(l) & 0.06 & 0.08 & 0.86 & 0.86 & -1.25 & 0.08 & \textbf{0.61} & 0.07 & \textbf{0.07} & 62.26 \\
GPT-2     & 0.05 & 0.12 & 0.82 & 0.86 & -1.34 & 0.36 & 0.57 & 0.21 & 0.12 & 55.06 \\
GPT-2-(m) & 0.05 & 0.12 & 0.83 & 0.86 & -1.35 & 0.35 & 0.56 & 0.22 & 0.14 & 52.91 \\
GPT-2-(l) & 0.05 & 0.12 & 0.83 & 0.86 & -1.34 & 0.35 & 0.55 & 0.21 & 0.16 & 52.88 \\
FlanT5-(s) & 0.06 & 0.12 & 0.83 & 0.84 & -1.35 & 0.31 & 0.57 & 0.26 & 0.12 & \textbf{73.82} \\
FlanT5-(b) & 0.06 & 0.11 & 0.84 & 0.84 & -1.34 & 0.29 & 0.51 & 0.22 & 0.16 & 70.51\\
FlanT5-(l) & 0.06 & 0.11 & 0.84 & 0.84 & -1.32 & 0.34 & 0.53 & 0.20 & 0.11 & 70.99\\
ChatGPT  & \textbf{0.08} & \textbf{0.17} & 0.67 & 0.81 & \textbf{-0.77} & \textbf{0.85} & 0.50 & \textbf{0.26} & 0.13 & 29.12 \\ \bottomrule
\end{tabular}
\caption{\footnotesize Evaluation of responses generated by each model for each counterspeech generation dataset in terms of generation, engagement and quality metrics. The first column denotes which model is being used for zero-shot evaluation. DialoGPT (DGPT) and GPT-2 has s, m and l suffixes which represent 117M, 345M and 762M parameter sizes, and FlanT5 has s, b and l suffixes which represent 80M, 250M and 750M parameter sizes. For evaluating generation we measure the average gleu, meteor (met), bleurt (blrt), novelty (nov) and diversity (div). Engagement metrics consist of upvote, width, and depth. For quality, we utilise the counterspeech (cs), argument (arg), counter argument (c\_arg) and toxicity (tox) scores and readilbility scores (fre). \textbf{Bold} denotes the best scores and higher scores denote better performance except for toxicity.}
\label{tab:vanilla-generation}
\end{table*}

\section{Evaluation metrics}
\label{sec:eval}
\noindent\textit{Generation metrics}: To measure the generation quality, we use different standard metrics. We use \texttt{gleu}~\cite{wu2016google} and \texttt{meteor}~\cite{banerjee2005meteor} to measure how similar the generated counterspeech are to the ground truth references.
We also measure if the LLMs generates diverse and novel counterspeech. For this purpose, we use the \texttt{diversity} and \texttt{novelty} metrics from existing literature~\cite{wang2018sentigan}. In addition, we also report one of the recent generation metrics, \texttt{bleurt}~\cite{sellam2020bleurt}. 
Note that, we do not use the BLEU~\cite{papineni2002bleu} score because it has some undesirable properties when used for single sentences, as it is designed to be a corpus-specific measure~\cite{wu2016google}. Further, the reader might notice negative scores in the case of \texttt{bleurt} metric. This is not unnatural since the \texttt{bleurt}, unlike BLEU, is not calibrated. For more information, refer here\footnote{https://github.com/google-research/bleurt/issues/1}. \\
\noindent\textit{Engagement prediction metrics}: We use the DialogRPT model~\cite{gao-etal-2020-dialogue} to predict the human feedback of the counterspeech generated using the following metrics -- \textit{width}: the number of direct replies to the given reply, \textit{depth}: the maximum length of dialog after this turn, and \textit{updown}: the number of upvotes minus the number of downvotes. 

This metric can help us in identifying how engaging the generated counterspeech is, which is another important characteristic, as noted by \citet{benesch2016considerations}. To calculate the engagement metric, we pass the \texttt{hate speech}-\texttt{counterspeech} pair to the model, which provides a score between 0 and 1 representing the engagement in terms of upvotes/width/height. This will denote the engagement probability of that metric for the given counterspeech. \\
\noindent\textit{Quality measurement metrics}: We deploy various third-party classifiers to evaluate the quality of the generated responses. To calculate the scores, we pass the generated counterspeech through the model and get the logit scores, which are passed through a softmax layer. The metrics used for evaluation are listed below.
\begin{compactitem}
    \item \textit{Argument}: In order to evaluate the argument characteristic of the generated response, we use a \texttt{roberta-base-uncased} model\footnote{https://huggingface.co/chkla/roberta-argument} fine-tuned  on the  argument dataset~\cite{stab-etal-2018-cross}. Given this model, we pass each generated response through the classifier to predict a confidence score, which would denote the argument quality.
    \item \textit{Counterargument}: In order to evaluate the counterargument  characteristic of the generated response, we use a \texttt{bert-base-uncased} model trained on the counterargument  dataset defined in section \ref{sec:additional}. We achieve an F1-score of $0.62$ on the test set of this dataset. Given this model, we pass each of the hate speech and the generated response through the classifier to predict a confidence score, which would denote the counterargument quality.
    \item \textit{Counterspeech}: In order to evaluate the counterspeech quality of the generated responses, we use a \texttt{bert-base-uncased} model trained on the counterspeech dataset introduced in section \ref{sec:additional}. We achieve an F1-score of $0.7$ on the test set of this dataset. Given this model, we pass each generated response through the classifier to predict a confidence score, which denotes the quality of the counterspeech.
    \item \textit{Toxicity}: We use the HateXplain model~\cite{mathew2020hatexplain} trained on two classes -- toxic and non-toxic\footnote{\footnotesize https://huggingface.co/Hate-speech-CNERG/bert-base-uncased-hatexplain-rationale-two}. We report the confidence between $[0,1]$ for the toxic class. This metric is important because a toxic counterspeech might escalate the discussion.
      
\end{compactitem}

\noindent\textit{Readability:} We further evaluate the readability of the counter speech generated. We use a popular readability metric - Fleish Reading Ease~\cite{farr1951simplification} (fre). It gives a score between 0-100.
\noindent\textit{Type classifier}: In order to evaluate the type specific generation, we train a \texttt{bert-base-uncased} model on the type based counterspeech data points mentioned in section~\ref{sec:additional} using a multi-class classification strategy. Overall, we achieve an average macro F1-score 0.80. Among the types, we achieve a macro F1-score $\sim$0.80 for denouncing, humor, facts and affiliation. Hypocrisy is hardest to classify with an F1-score of 0.59 and questions are the easiest to classify with an F1-score of 0.97.

\section{Results}

\subsection{Vanilla generation}

Here we discuss the evaluation results for the zero-shot evaluation of various models for the vanilla generation setting. 

\noindent\textbf{Does counterspeech generation depend on model size in zero-shot setting?} We compare the small, medium (base) and large sizes of three different variations of models, i.e., DialoGPT, FlanT5 and GPT-2. We note the percentage change between the largest and smallest model. We observe that for the synthetic datasets, i.e., CONAN and CONAN-MT the change is not significant in terms of generation metrics. In terms of counterspeech quality, we see a drop of 42\%  for the DialoGPT model for the CONAN dataset, whereas there is a drop of 13\% for GPT-2 in case of CONAN-MT dataset. In terms of counterargument quality, we notice a drop of 6-9\% across all the models except DialoGPT for CONAN dataset. Surprisingly, for CONAN-MT dataset, the toxicity increases by 44\% for the GPT-2 models as we increase the model size. For the real world datasets, i.e., Gab and Reddit, there is a significant increase in toxicity (25-30\%) as we increase the size of GPT-2 model. On the other hand, we find that there is an improvement in the generation quality for the DialoGPT in terms of gleu and meteor metrics. In addition, we find the readability of the counterspeech generated by DialoGPT on the Reddit dataset increases dramatically (100\%) with increase in the size of the model. The size trend for Flan-T5 is not consistent although it performs quite poorly compared to DialogGPT and GPT-2. \\
\noindent\textbf{Does counterspeech generation depend on model type in zero-shot setting?} We compare DialoGPT, GPT-2 and FlanT5 models since they vary in their architecture, pretraining (finetuning) strategies and dataset used for pre-training. In terms of synthetic datasets, FlanT5 models are better in terms of \texttt{gleu} (30\%) and meteor (50\%) and GPT-2 models are better in \texttt{bleurt} (8-15\%). In terms of other metrics, we find that counterspeech quality is far better for GPT-2 models (200\%) than other models. DialoGPT models are also better in terms of readability, but at the same time are more toxic.  For the real world datasets, i.e., Gab and Reddit, DialoGPT models are better in terms of \texttt{bleurt} (7-10\%), GPT-2 models are better in meteor and FlanT5 models are better in terms of \texttt{gleu} (20\%).  In terms of other metrics, we find that counterspeech quality and argument improves by 2–3 times for GPT-2 and FlanT5 models, but they are also higher in terms of toxicity.

\begin{table}[!htpb]
\centering
\scriptsize
\begin{tabular}{lp{2cm}p{2cm}}\toprule
\textbf{Model} & \textbf{Hate speech} & \textbf{Counterspeech} \\\midrule
FlanT5 & Muslims have no place in civilised Europe. They should go home. & We shall not discriminate on the grounds of race, religion, colour, culture or national origin. Our society defends and protects its own dignity. \\\midrule
DGPT & We are being invaded by Muslims. They are taking over our countries & lol s I hope we all get this reference.The irony is lost at the end of that sentence isn't it...\\\midrule
ChatGPT & If you are a Jew you are for sure involved in 9/11 planning. This is all about Race and 'Enlightenment'. You are all guilty until proven innocent & It is unfair and untrue to suggest that all Jews were involved in the planning of 9/11... \\\bottomrule
\end{tabular}
\caption{\footnotesize Counterspeech generation by different models in a zero-shot setting for vanilla generation. The posts are taken from one of the available datasets. For all models except ChatGPT, we use the \texttt{l} version of the models.}
\label{tab:table-example-cs1}
\end{table}

\begin{table}
\scriptsize
\begin{tabular}{l|l|c|c|c|c|c|c}\toprule
\textbf{model}&\textbf{prompt}&\textbf{aff}&\textbf{den}&\textbf{fac}&\textbf{hum}&\textbf{hyp}&\textbf{qu}\\\midrule
\multicolumn{8}{c}{\textbf{CONAN\_MT}} \\\midrule
\multirow{4}{*}{GPT-2}      & base & 0.04 & 0.04 & 0.60 & 0.03 & 0.29 & 0.00 \\
                            & manual & 0.06 & \textbf{0.40} & \textbf{0.78} & \textbf{0.20} & 0.21 & 0.01 \\
                            & freq & 0.06 & 0.07 & 0.77 & 0.10 & 0.21 & \textbf{0.05} \\
                            & cluster & \textbf{0.46} & 0.34 & 0.72 & 0.07 & \textbf{0.43} & 0.03 \\\midrule
\multirow{4}{*}{DGPT}       & base & 0.04 & 0.10 & 0.29 & 0.19 & 0.39 & 0.00 \\
                            & manual & 0.10 & \textbf{0.45} & \textbf{0.46} & \textbf{0.46} & \textbf{0.51} & 0.01 \\
                            & freq & 0.10 & 0.15 & 0.44 & 0.34 & 0.50 & \textbf{0.08} \\
                            & cluster & \textbf{0.38} & 0.41 & 0.42 & 0.29 & 0.49 & 0.03 \\\midrule
\multirow{4}{*}{FlanT5}     & base & 0.04 & 0.06 & 0.73 & 0.03 & 0.14 & 0.00 \\
                            & manual & 0.15 & \textbf{0.23} & \textbf{0.81} & \textbf{0.15} & 0.10	& 0.00 \\
                            & freq & \textbf{0.26} & 0.06 & 0.80 & 0.07 & 0.21 & 0.00 \\
                            & cluster & 0.20 & 0.18 & 0.74 & 0.10 & \textbf{0.23} & 0.00 \\\midrule                            
\multirow{4}{*}{ChatGPT}    & base & 0.02 & 0.22 & 0.75 & 0.00 & 0.01 & 0.00 \\
                            & manual & 0.03 & 0.40 & 0.81 & 0.00 & 0.02 & 0.00 \\
                            & freq & \textbf{0.51} & 0.31 & \textbf{0.94} & 0.00 & 0.06 & 0.01 \\
                            & cluster & 0.27 & \textbf{0.46} & 0.77 & 0.00 & \textbf{0.09} & \textbf{0.01} \\\midrule
\multicolumn{8}{c}{\textbf{CONAN}} \\\midrule
\multirow{4}{*}{GPT-2}       & base & 0.02 & 0.04 & 0.65 & 0.01 & 0.27 & 0.00 \\
                            & manual & 0.05 & \textbf{0.37} & \textbf{0.81} & \textbf{0.13} & 0.19 & 0.01 \\
                            & freq & 0.04 & 0.07 & 0.79 & 0.05 & 0.18 & \textbf{0.04} \\
                            & cluster & \textbf{0.44} & 0.33 & 0.76 & 0.04 & \textbf{0.39} & 0.02 \\\midrule
\multirow{4}{*}{DGPT} & base & 0.02 & 0.09 & 0.28 & 0.20 & 0.41 & 0.00 \\
                            & manual & 0.07 & \textbf{0.44} & \textbf{0.42} & \textbf{0.46} & \textbf{0.52} & 0.01 \\
                            & freq & 0.07 & 0.16 & 0.42 & 0.37 & 0.50 & \textbf{0.07} \\
                            & cluster & \textbf{0.36} & 0.41 & 0.35 & 0.32 & 0.50 & 0.03 \\\midrule
\multirow{4}{*}{FlanT5} & base & 0.02 & 0.08 & 0.75 & 0.02 & 0.12 & 0.00 \\
                            & manual & 0.10 & \textbf{0.23} & 0.79 & \textbf{0.14} & 0.12 & 0.00 \\
                            & freq & \textbf{0.18} & 0.07 & \textbf{0.80}& 0.05 & 0.21 & 0.00\\
                            & cluster & 0.16 & 0.17 & 0.73 & 0.08 & \textbf{0.23} & 0.00 \\\midrule
\multirow{4}{*}{ChatGPT}    & base & 0.04 & \textbf{0.64} & 0.28 & 0.00 & \textbf{0.10} & 0.00 \\
                            & manual & 0.02 & 0.39 & 0.93 & 0.00 & 0.02 & 0.00 \\
                            & freq & \textbf{0.52} & 0.30 & \textbf{0.96} & 0.00 & 0.06 & 0.00 \\
                            & cluster & 0.26 & 0.43 & 0.87 & \textbf{0.01} & 0.10 & \textbf{0.01} \\\midrule
\multicolumn{8}{c}{\textbf{Reddit}} \\\midrule
\multirow{4}{*}{GPT-2}       & base & 0.08 & 0.07 & 0.27 & 0.17 & 0.41 & 0.00 \\
                            & manual & 0.20 & \textbf{0.48} & \textbf{0.51} & \textbf{0.40} & 0.48 & 0.01 \\
                            & freq & 0.20 & 0.16 & 0.50 & 0.26 & 0.50 & \textbf{0.08} \\
                            & cluster & \textbf{0.47} & 0.41 & 0.47 & 0.21 & \textbf{0.50} & 0.03 \\\midrule
\multirow{4}{*}{DGPT}       & base & 0.03 & 0.07 & 0.14 & 0.47 & 0.30 & 0.00 \\
                            & manual & 0.08 & \textbf{0.43} & \textbf{0.30} & \textbf{0.75} & \textbf{0.52} & 0.01 \\
                            & freq & 0.08 & 0.14 & 0.29 & 0.64 & 0.52 & \textbf{0.07} \\
                            & cluster & \textbf{0.33} & 0.38 & 0.29 & 0.52 & 0.41 & 0.02 \\\midrule
\multirow{4}{*}{FlanT5}     & base & 0.09 &	0.12 & 0.26 & 0.23 & 0.30 & 0.00 \\
                            & manual & 0.17	& \textbf{0.27} & \textbf{0.44} & \textbf{0.34} & 0.30 & 0.00 \\
                            & freq & \textbf{0.23} & 0.12 & 0.41 & 0.25 & 0.30 & \textbf{0.01} \\
                            & cluster & 0.19 & 0.21 & 0.43 & 0.24 & \textbf{0.31} & \textbf{0.01}\\\hline
\multirow{4}{*}{ChatGPT}    & base & 0.07 & 0.39 & 0.44 & 0.00 & 0.10 & 0.00 \\
                            & manual & 0.07 & \textbf{0.60} & \textbf{0.75} & 0.02 & 0.10 & 0.00 \\
                            & freq & \textbf{0.54} & 0.48 & 0.75 & 0.00 & \textbf{0.13} & 0.01 \\
                            & cluster & 0.35 & 0.57 & 0.57 & 0.01 & 0.12 & \textbf{0.01} \\\midrule
\multicolumn{8}{c}{\textbf{Gab}} \\\midrule
\multirow{4}{*}{GPT-2}      & base & 0.08 & 0.09 & 0.26 & 0.18 & 0.40 & 0.00 \\
                            & manual & 0.23 &\textbf{0.49} & \textbf{0.51} & \textbf{0.40} & 0.47 & 0.01 \\
                            & freq & 0.23 & 0.19 & 0.48 & 0.28 & 0.46 & \textbf{0.07} \\
                            & cluster & \textbf{0.50} & 0.44 & 0.45 & 0.20 & \textbf{0.49} & 0.03 \\\midrule
\multirow{4}{*}{DGPT}       & base & 0.03 & 0.08 & 0.12 & 0.47 & 0.29 & 0.00 \\
                            & manual & 0.09 & \textbf{0.46} & \textbf{0.27} & \textbf{0.73} & 0.52 & 0.01 \\
                            & freq & 0.09 & 0.16 & 0.27 & 0.64 & \textbf{0.53} & \textbf{0.07} \\
                            & cluster & \textbf{0.33} & 0.41 & 0.26 & 0.52 & 0.39 & 0.02 \\\midrule
\multirow{4}{*}{FlanT5}     & base & 0.10 &	0.12 & 0.24 & 0.25 & 0.29 & 0.00 \\
                            & manual & 0.17 & \textbf{0.28} & \textbf{0.43} & \textbf{0.37} & 0.29 & 0.00 \\
                            & freq & \textbf{0.23} & 0.11 & 0.41 & 0.28 & 0.29 & 0.00 \\
                            & cluster & 0.19 & 0.23 & \textbf{0.43} & 0.26 & \textbf{0.33} & 0.00\\\midrule
\multirow{4}{*}{ChatGPT}    & base & 0.02 & 0.09 & \textbf{0.88}& 0.00 & 0.01 & 0.00 \\
                            & manual & 0.07 & \textbf{0.71} & 0.62 & \textbf{0.01} & 0.07 & 0.00 \\
                            & freq & \textbf{0.52} & 0.66 & 0.65 & 0.00 & \textbf{0.09} & 0.00 \\
                            & cluster & 0.30 & 0.69 & 0.52 & 0.00 & 0.07 & 0.00 \\\bottomrule
\end{tabular}
\caption{\scriptsize Evaluation of responses generated by each model and prompt strategy. The first column denotes which model is being used for zero-shot evaluation. DialoGPT, GPT-2 and FlanT5 are averaged across three parameter sizes. The second column denotes the prompt strategy out of manual\_prompt (manual), frequency based (freq), cluster centered (cluster) being used, where baseline (base) represents no prompt strategy. The next six columns represent the type-precision for the counterspeech generated by each model + \texttt{type\_prompt}. aff: affiliation, den: denouncing, fac: facts, hum: humour, hyp: hypocrisy, qu: question. \textbf{Bold} denotes the best scores and higher scores denote better performance.}
\label{tab:type-specific-generation}
\end{table}

\subsection{Type specific generation} In this part, we evaluate the type specific generation using type prompts. We run the counterspeech classifier (described in section \ref{sec:eval}) over the post generated for a particular type and measure the ratio of posts which the classifier classifies as the same type. We name this metric as \texttt{type precision}. The type prompts and their extraction procedure is mentioned in section \ref{sec:prompt}. Since we do not observe much change in the type specific generation for small, medium and large versions of the DialoGPT, FlanT5 GPT-2, we present their average performance in the Table~\ref{tab:type-specific-generation}. 
For \textit{affiliation}, GPT-2 and DialoGPT perform better with cluster centered prompts across all the datasets, while FlanT5 performs better with frequency based prompts. Cluster centered prompts improve the baseline type precision by 0.41 and 0.32 units for GPT-2 and DialoGPT respectively and frequency based prompts improve the FlanT5 by 0.16 units. For \textit{denouncing}, GPT-2, DialoGPT and FlanT5 perform better with manual prompts. The type precision of FlanT5 increases by 0.16 units while the other models increases by 0.36 units. \textit{Facts} type\footnote{We don't claim that these models are generating factual statements, but fact-styled statements.} are one of the categories where the GPT-2 and 
FlanT5 performs quite better than DialoGPT type precision. Overall, the baseline FlanT5 model is also good for generating fact type statements for CONAN and CONAN-MT datasets, with scoring $\sim$0.7 in type precision. Here, also manual prompts are the best, and they improve the baseline type precision by 0.20, 0.15 and 0.13 units for GPT-2, DialoGPT and FlanT5 respectively. For \textit{humour}, we find again that the manual prompts are the best. They improve the baseline type precision by 0.185, 0.27 and 0.12 units for GPT-2, DialoGPT and FlanT5 respectively. For \textit{hypocrisy} and \textit{questions}, there is no one prompting strategy that works for both the models and all the datasets. Cluster centered prompts are better for GPT-2 (0.10 units) and manual prompts are better for DialoGPT (0.15 units). Surprisingly, none of the models reach 0.1 type precision for the \textit{question} type. We have added some examples of type specific generation in the Table~\ref{tab:table-example-cs2}.

\begin{table}[!t]
\centering
\scriptsize
\begin{tabular}{lp{2cm}p{2cm}}\toprule
\textbf{Model (Type)}  & \textbf{Hate speech} & \textbf{Counterspeech} \\\midrule
GPT-2 (aff)  & Blacks are just an inferior bloodline to our own. We should remove them. &  i am muslim but my black brothers don't respect us as true Muslims, and there is another cause of that.. so how do I make their hearts be happy then they can accept me? the one thing we have in common with other races....is faith - you know all over this world...every people has a belief" \\\midrule
GPT-2 (den) &  No, only the Severely Retarded get the job. & i think maybe prejudice against people with that disease does drive a lot of discrimination in their lives - they've got some serious issues and it's very hard to help them \\\midrule
DialoGPT (hum) &  Never said it was, faggot & This is funny. uhh you know who loves the attention and self hate speech as well? Donald Trump, \\\midrule
DialoGPT (hyp) & People of a different race, including transgenders, are very few. With some effort we can contain them. &  In contradiction to the people living in fear and ignorance. gender neutrality exists even though those who want it aren\'t happy about their pronouns? \\\midrule
FlanT5 (aff) & We should expel Muslims, everyone wants them out of the country. & I have no issue with Muslims. I just disagree that they are the bad guys and there are lots of other countries like my country in which they won’t go, so we shouldn’t do anything about them. \\\midrule
ChatGPT (facts) & Women should not be allowed to vote. They are subhuman. & I am a Muslim and in our faith, both men and women are valued equally. Discriminating against one gender is unjustified, as it goes against the principles of justice that Islam teaches us to follow .. \\\bottomrule
\end{tabular}
\caption{\footnotesize Counterspeech generated by different models in a zero-shot setting for type specific generation where the type is denoted in the brackets. For each type, we select the prompt strategy based on the performance of that model on that type. aff: affiliation, den: denouncing, fac: facts, hum: humour, hyp: hypocrisy, qu: question. For all models except ChatGPT, we use their larger versions.}
\label{tab:table-example-cs2}
\end{table}

\subsection{Generation using ChatGPT}

In this part, we look at the counterspeech generated by ChatGPT. In terms of vanilla generation, as noted in Table \ref{tab:vanilla-generation}, we notice that ChatGPT performs better than other models in terms of generation quality. It improves the gleu (12\%), meteor (32\%) and bleurt (42.25\%). Among other metrics, ChatGPT improves the counterspeech quality by 120\% and improves the argument quality by 27\%. The toxicity scores are comparable, although they are slightly higher than the best models. One interesting point is that readability of the ChatGPT texts reduces by $\sim$35\%. We also note some counterspeech generated by ChatGPT and compare them with the other models in Table~\ref{tab:table-example-cs1}. In terms of type specific generation as noted in Table~\ref{tab:type-specific-generation}, we find that for \textit{affiliation}, frequency based prompts improves the baseline type precision by 0.49 units across all datasets. Other than that, other types do not have a consistent best prompt strategy that works across all the categories. Interestingly, for some cases like denouncing type for CONAN, the model performance worsens if we introduce any prompt. Further ChatGPT model performs best in fact-type counterspeech scoring close to 0.9 in three out of four datasets. ChatGPT struggles for the types—humour, hypocrisy, and questions even in presence of type prompts and their type precisions rarely reach above 0.1 across datasets.

\section{Discussion and conclusion}

In this work, we presented a thorough analysis of the performance of LLMs in a zero-shot setting to generate counterspeech. This is to understand what these models are capable of intrinsically without training on hate speech-counterspeech pairs. We explored the models namely DialoGPT, GPT-2 and FlanT5 in terms of their size and further extended the experiments to ChatGPT. 

In case of vanilla experiments, we find these models show some promise in generating counterspeech in zero-shot setting, with ChatGPT outperforming the other two models. Further, we do not see many changes in the variations among the GPT-2 models. The improvement for ChatGPT is also visible as we manually evaluated the generation (shown in Table ~\ref{tab:table-example-cs1}). Hence, like other tasks, the emergent behaviour is only visible when we increase the scale (i.e., GPT-2 $\rightarrow$ ChatGPT).

Next, we proposed three prompting strategies to generate categorical counterspeech and analysed the applicability of all these different models for the same. We find that carefully designed manual prompts are better than our proposed automatic methods. Although, these prompts are able to control the generations for a smaller model like DialoGPT, it fails for ChatGPT except for some specific types like facts. This opens the door for future research, which can focus on two directions (a) design of better prompting strategies, and (b) improve models like ChatGPT to better capitalise on these prompts to benefit the type specific generations for LLMs.

\noindent\textbf{Do the metrics correlate with human judgements?} While we present most of our results with automatic metrics, it is important to understand if they correlate with human judgements. We took one referential (\texttt{bleurt}) and non referential metric (counterspeech). For each metric, 25 samples were extracted from each tail of the predicted metric values. We present these to expert researchers in the hate speech domain and ask them to rate the quality of counterspeech from 1-5, 5 being the best and 1 being the worst. For the counterspeech metric (nominal ratings, ordinal human evaluations), the Point-biserial correlation coefficient~\cite{linacre2008expected} was 0.45. For the bleurt metric (continuous ratings, ordinal human evaluations), the Spearman's rank correlation was 0.56. These results highlight the consistency between automated metrics and human judgments, affirming their reliability.

\section*{Ethics statement}
Hate speech is a complex phenomenon. While the language generation methods are better than before, it is still very far from generating coherent and meaningful counterspeech~\cite{10.1145/3442188.3445922}. Further, they are very unreliable. 
There are multiple cases where the chatbots turned hateful when deployed without supervision, leading to their shutting down\footnote{https://www.cbsnews.com/news/microsoft-shuts-down-ai-chatbot-after-it-turned-into-racist-nazi/}.  Hence, we advocate against the deployment of fully automatic pipelines for countering hate speech~\cite{delosRiscos2021}. Based on the current progress in this pipeline, active participation of the counter speakers is required to generate relevant counterspeech. Our efforts to study the counterspeech generation by these automatic models can help in further improvement in the counterspeech generation pipeline and better inform the counter speakers.

\section{Bibliographical References}\label{sec:reference}

\bibliographystyle{lrec-coling2024-natbib}
\bibliography{lrec-coling2024-example}

\end{document}